# Diffusion Reinforcement Learning Based Online 3D Bin Packing Spatial Strategy Optimization

Jie Han, Tong Li, Qingyang Xu, Yong Song, Bao Pang, and Xianfeng Yuan

*Abstract*—The online 3D bin packing problem carries important research value in logistics, warehousing and intelligent manufacturing. Solutions to this problem have gradually shifted from traditional heuristic algorithms to deep reinforcement learning-based approaches. However, deep reinforcement learning faces challenges such as low sample efficiency, high dimensionality, and unstable policy optimization in online 3D bin packing tasks, which limit the algorithm's generalization and adaptability to items of various sizes. This paper proposes an online 3D bin packing algorithm based on diffusion reinforcement learning. The algorithm models the packing process as a Markov decision chain and introduces a height map-based state representation to improve the agent's environmental perception. In the policy generation stage of the AC algorithm, a diffusion model-based actor network is adopted to model complex action distributions. The actor network outputs action probability distributions and guides the agent to generate more reasonable packing actions, thereby improving policy convergence speed and space utilization. Experimental results on multiple public datasets show that, compared with state-of-the-art reinforcement learning-based 3D bin packing methods, the proposed algorithm achieves a significant improvement in the average number of packed items, demonstrating excellent application potential in complex online decision-making scenarios..

*Index Terms*—Online 3D Bin Packing, Diffusion Model, Reinforcement Learning, Action Generation

## I. Introduction

IN logistics, warehousing, and manufacturing, various items are efficiently loaded into limited space to enhance resource utilization and system efficiency. As a key research direction in spatial optimization, the Bin Packing Problem (BPP) is widely studied in operations research and artificial intelligence fields. In three-dimensional space, the 3D Bin Packing Problem (3D-BPP)[1] is recognized as a classic combinatorial optimization issue due to its alignment with real-world needs. It is increasingly applied in multiple industrial fields. The core objective is to place items reasonably into containers with limited volume, with container and item sizes known, to maximize space utilization.

The traditional three-dimensional BPP generally assumes that all item information is fully known before decision-making begins, allowing for the application of global optimization algorithms to obtain optimal or near-optimal solutions. However, in many real-world scenarios, such as dynamic project management, real-time transportation scheduling, or online order sorting and selection, item information often arrives gradually over time and is difficult to acquire in advance. Against this backdrop, the online three-dimensional BPP has emerged. This problem requires a robotic arm to make packing decisions in a timely manner based on the current packing state, as items arrive one by one and their information is revealed step by step. This situation places higher demands on the real-time performance of algorithms, the quality of decisions, and the robustness of the system. Compared with the traditional three-dimensional BPP, the online version faces more complex challenges, which are mainly reflected in the following three aspects: (1) Incomplete information: In the online 3D-BPP, item sizes, shapes, and packing sequences are unknown before arrival. Algorithms cannot plan in advance like traditional methods. Decisions must be made based on real-time arriving items, increasing uncertainty. (2) High dimensional action space: Each item's placement position and orientation offer multiple choices, resulting in a large action space. For large-scale problems, efficiently exploring this high dimensional space without increasing computational complexity is challenging. (3) Real-time requirements: Decisions must be made immediately upon item arrival. Algorithms must have high real-time responsiveness. Compared with offline problems, online algorithms must address packing quality and complete decisions in a short time.

With the rapid development of e-commerce, intelligent warehousing, and automated production lines, the online 3D BPP is increasingly recognized as a critical technical challenge. As a Non-deterministic Polynomial problem, traditional heuristic algorithms, which rely on greedy objectives, are designed by mimicking human packing experience. Although intuitive, optimal solutions are rarely achieved. In recent years, deep reinforcement learning (DRL) has emerged as a promising strategy for addressing the online 3D-BPP due to its effective decision-making capabilities in complex dynamic environments.

In recent years, DRL has been widely applied to the online 3D-BPP due to its ability to interact and optimize strategies in dynamic environments[2], [3], [4], [5]. However, it still faces challenges such as low sample efficiency, high dimensional state-action spaces, and unstable policy optimization. These issues significantly limit DRL's generalization in BPPs. Trained models cannot be transferred to scenarios with diverse item types. Moreover, existing methods are typically limited to inferring items of the same size as those used in training[3], [6]. Transferring trained packing policies to items of different sizes remains difficult. Additionally, varying item sizes greatly affect the action space size and the probability distribution of selecting appropriate placement actions, especially when large size disparities are handled in real time.

To address these issues, an online 3D bin packing method based on diffusion reinforcement learning is proposed. In this method, a placement generation module first generates a fixed-length set of free subspaces as placement candidates based on the current feature vector. Consistency in the size of the packing action space is ensured. Placement candidates and packing items are jointly defined as the state of a Markov decision process. Then, a diffusion model is utilized to model complex multimodal action distributions effectively through a progressive denoising process of data distribution. Action distributions matching the characteristics of sample data are generated. The difficulty of exploring high dimensional action spaces faced by traditional policies is alleviated. Policy quality and learning efficiency are improved. When integrated into a reinforcement learning framework, exploration efficiency is enhanced. More reasonable and diverse packing decisions are generated in complex, high dimensional 3D bin packing environments. A new technical approach for solving the online 3D-BPP is provided. The main contributions of this paper are as follows:

1) Markov modeling and height map state construction for online bin packing: The online 3D-BPP (Online 3D-BPP) is formalized as a Markov decision process (MDP). Environment observations are constructed using height maps based on the current container state to enhance structural information capture capability.

2) Diffusion model-based action network: A diffusion model is introduced as a policy generator to model the packing action probability distributions which improves the policy quality and learning efficiency.

3) The proposed method is trained and tested on several typical bin packing datasets (e.g., RS, CUT-1, CUT-2). Experimental results show significant improvements in space utilization and increased efficiency in the time consuming for decision-making.

## II. Related Work

### A. Heuristic Methods

The online 3D-BPP is recognized as a classic optimization challenge and is classified as NP-hard[4]. Infinite possible solutions exist, and verifying each solution's correctness in polynomial time[7] is not feasible. Feasible solutions have been proposed by many researchers to address this issue. Early studies on 3D-BPP focused on rules derived from human workers' experience. Quick feasible or near-optimal solutions were found using methods like tabu search[8], first-fit algorithms[9], and heuristic algorithms[10]. An online bin packing heuristic algorithm (Online BPH) was proposed by Ha et al.[11], to tackle the 3D container loading problem (3D-CLP) in dynamic environments. The algorithm was designed for online scenarios, enabling real-time processing of arriving items. Item placement was optimized by introducing the concept of empty maximum space (EMS), preventing item blockages. A new 3D positioning heuristic method was proposed by Wang et al.[12]. The search process was accelerated using heightmaps. A basic heuristic block loading algorithm was introduced by Zhang et al.[13]. A block determined by a block selection algorithm was loaded in each packing stage following a fixed strategy until no blocks were available. A multilayer search algorithm based on depth-first search was developed to select blocks for each packing stage, achieving results closer to the optimal solution. Manually designed packing methods were often inefficient and lacked intelligence. Approximate optimal solutions were sought by some researchers using intelligent optimization algorithms, such as greedy algorithms, simulated annealing, genetic algorithms, and differential evolution algorithms. A greedy search heuristic algorithm was proposed by Silva et al.[14], to solve 3D-BPP. Maximum space utilization and box stability were required. A mixed integer linear programming model was developed by Mostaghimi et al.[15], to address loading rectangular boxes into containers. Total packing value was maximized. A sequence triplet-based solution method was proposed, and simulated annealing was used as a modeling technique to study cases where some boxes were pre-placed in containers.

### B. DRL-based Methods

The application of genetic algorithms in an intelligent packing simulator for 3D-BPP was studied by Khairuddin et al.[16]. An adaptive chromosome length genetic algorithm was adopted by the intelligent packing system. Box movement and positioning were simulated through GA iterations to achieve realistic adaptability. Boxes were optimally arranged to fit into the smallest possible containers. A new technique combining differential evolution with a trinary search tree model (TSTDE) was proposed by Huang et al.[17], to solve 3D-CLP. A set of suboptimal solutions was generated using the trinary tree model as the initial population for DE. An improved differential evolution algorithm was then used to search for feasible solutions. Although these methods offer some intuitiveness and effectiveness, reliance on manually designed rules is heavy. Consistent superior performance across different problem scenarios is difficult to achieve. In contrast, excellent performance in solving certain combinatorial optimization problems is demonstrated by DRL[18]. The limitations of traditional domain expert knowledge are overcome by DRL techniques in addressing BPPs. More flexible and efficient strategies are provided. Better optimization strategies have been explored by some researchers using DRL-based methods for 3D-BPP[19], [20]. A data-driven tree search algorithm (DDTS) was proposed by Zhu et al.[21], to handle 3D-BPP. The solution space with complex constraints was explored using a tree search algorithm. A convolutional neural network trained on historical data was used to guide tree pruning, accelerating the search process. A novel brain-inspired experiential reinforcement model was proposed by Zhang et al.[22], imitating human experience-based reasoning. Advantages of biological and engineering systems were integrated. Experience from similar scenarios was learned, enabling adaptation to complex scenes and changing environments like the human brain. A Transformer-based DRL model was proposed by Que et al.[23], to address offline 3D-BPP with variable heights. However, real-time strategy adjustments for unexpected situations are challenging in offline 3D-BPP. In contrast, the online BPP addressed in this paper holds greater practical research significance. DRL was combined with Monte Carlo tree search (MCTS) by Jia et al.[24]. Information about future items was utilized to assist decision-making. The convergence issue in high dimensional discrete action spaces was alleviated through an improved Actor-Critic algorithm and a heuristic-based packing configuration tree model. A constrained DRL framework was built by Zhao et al.[2], on this basis. A feasibility mask prediction module was introduced to guide the agent to focus on feasible actions during training. Learning efficiency and policy quality were improved. A packing configuration tree (PCT) was proposed in their subsequent work[5] as a structured representation of packing states and action spaces. The DRL policy learning process was further simplified, performing well even in continuous solution spaces. Heuristic rules have been gradually integrated with reinforcement learning in existing studies. The integration of physical, packing, and unpacking heuristic methods into a DRL framework was systematically explored by Yang et al.[6]. A heuristic-guided DRL model was constructed, significantly enhancing the practicality and generalization of online 3D-BPP strategies. A candidate graph generated by heuristics was introduced by Xiong et al.[3], to identify potentially feasible placement positions. Efficient searches in large discrete action spaces were achieved. The model was successfully transferred to real robotic systems, verifying its application value. The unpacking mechanism was incorporated into a DRL framework by Song et al.[25]. Synergy between packing and unpacking behaviors was learned. Incorrectly packed items could be corrected by unpacking, improving the utilization rate of packing strategies. An adjustable robust reinforcement learning (AR2L) framework was proposed by Pan et al.[26]. Robustness weights were efficiently adjusted to achieve an ideal balance between average and worst-case policy performance. A hierarchical tree A2C (Advantage Actor-Critic) reinforcement learning framework was proposed by Zhang et al.[27], to address online air luggage packing. A complex packing task was first decomposed into smaller, manageable subtasks. Independent optimization was performed at each layer using the A2C algorithm. A top-down optimization algorithm was formed, with low-level optimization details feeding back to adjust higher-level strategies, ultimately achieving global optimization. A model based on "chunk-based dual-task learning" was proposed by Liu et al.[28]. Deep learning techniques were used to systematically learn two related tasks: determining the optimal orientation and position of each item in the container. Performance on both tasks was improved by leveraging commonalities between related tasks, enabling better generalization across different packing scenarios. Zhao et al.[29], note that existing neural combinatorial optimization methods for the 3D-BPP suffer from suboptimal performance due to diverse observations and sparse rewards. They propose the DMRL-BPP framework, incorporating a dynamic multi-modal encoder and a novel reward function to address the sparse reward issue. Zhang et al.[30], proposed a heuristic-guided DRL framework for solving online BPP. This approach employs an inheritance tree structure to represent states and utilizes advanced value-based DRL algorithms to train the packing agent. Furthermore, the researchers developed a novel data augmentation technique tailored for BPP, effectively accelerating training processes while reducing positioning operations.

## III. PROBLEM DESCRIPTION

To simplify the problem, the online BPP is defined as follows: Items arrive cyclically, and complete information about all items cannot be obtained beforehand. Rectangular items with random dimensions $l_i$, $w_i$, $h_i$ are placed by an agent into a container with dimensions $L$, $W$, $H$. The container's space utilization is maximized. To ensure stable stacking of items, seven constraints must be met, including:

(1) Support constraint: Stability during stacking and transportation is improved by ensuring full or partial effective support for the item's base. To accelerate the learning process and experience accumulation in intelligent systems, a partial support constraint is applied. The item's center of mass must be positioned above the vertical projection of the support area. Partial areas are permitted to remain in a non-contact floating state. Training efficiency and effectiveness are enhanced.

(2) Orientation constraint: To ensure standardized and stable placement during stacking, items are required to adhere to six predefined placement orientations. Specific requirements are illustrated in Figure 1.

(3) Boundary constraint: Items must not exceed the container's boundaries.

(4) Parallel constraint: During stacking, each layer must be kept strictly parallel to the container's inner walls. Neatness and stability of the stacking structure are ensured.

(5) Non-overlap constraint: To utilize container space effectively and prevent interference or damage, items must not overlap within the container.

(6) Time constraint: Due to the online bin packing nature, optimal stacking decisions must be made by the agent within 0.01 seconds.

(7) Loading order constraint: During loading, items must be arranged following the principle of larger and heavier items at the bottom and smaller and lighter items on top.

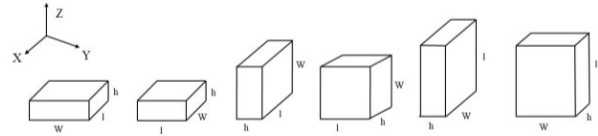

Fig. 1. Schematic diagram of item placement orientation.

Under the seven constraints mentioned above, the core optimization goal is to maximize the utilization of the container's internal space. The specific quantitative metric, space utilization rate $S_p$, is calculated using the formula provided below:

$$S_p = \frac{\sum_{i=1}^{N} l_i \cdot w_i \cdot h_i}{L \cdot W \cdot H} \times 100\%, i \in \{1,2,3,\ldots,N\} \tag{1}$$

A Cartesian 3D coordinate system is used to define the positions of the rectangular container and objects in 3D space. As shown in Figure 2, the origin is set at the front-left-bottom corner of the rectangular container. A 3D rectangular coordinate system is established with the container's length, width, and height corresponding to the x-axis, y-axis, and z-axis, respectively. To specify the spatial arrangement of items in the container, the corner coordinate method is adopted. The front-left-bottom corner point of an item is used as the reference point.

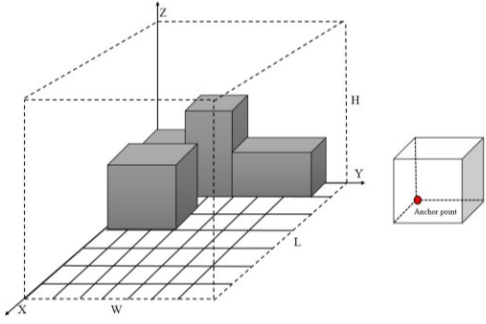

Fig 2. Schematic diagram of cargo corner coordinates.

## IV. METHOD

The overall architecture of the online 3D bin packing algorithm based on the diffusion reinforcement learning (DRL) framework is shown in Figure 3. A closed-loop system comprising state perception, policy generation, action filtering, and environment interaction is constructed to optimize packing decisions in complex dynamic environments.

Specifically, a height map is first constructed based on the current container state. Geometric information of the items to be packed is combined to form a complete state input. Structural features are extracted from this state information using a convolutional neural network (CNN). These features are then fed into the policy network and value network within an Actor-Critic structure. Candidate action distributions are generated by the policy network. Action values are evaluated by the value network.

To enhance flexibility and distribution modeling in action generation, a diffusion model is introduced. Complex multi-modal distributions in high dimensional action spaces are modeled, improving the policy's expressiveness for nonlinear packing scenarios. A feasibility mask predictor module is incorporated to predict valid action sets based on the current state. Action distribution constraints and pruning are achieved, ensuring generated policies satisfy physical constraints like space, support, and boundaries.

Finally, interactions are executed by the agent based on filtered actions, and corresponding reward feedback is received. Interaction data are stored in an experience replay pool for subsequent policy updates. The diffusion model, policy network, and value network are synchronously optimized under a joint training mechanism. Policy convergence and generalization in high dimensional dynamic environments are improved.

### A. Actor and Critic Network Design

Reinforcement learning has made remarkable achievements in solving complex decision-making problems. In 3D bin pac-king tasks, reinforcement learning can learn the placement positions and attitudes of objects. Through trial and error, the agent can continuously optimize packing strategies and achieve high space utilization. In this research, to address the challenges of high dimensional action spaces, multiple constraints, and sparse feedback in packing tasks, we construct a learning architecture based on the Actor-Critic framework. We combine state encoding networks with feasibility-constrained modeling to improve the efficiency and rationality of strategy generation. The Actor-Critic algorithm consists of a policy network (Actor) and a value network (Critic). The Actor network interacts with the environment and outputs action probability distributions. The Critic network evaluates the state values of the actions selected by the Actor and guides policy optimization through backpropagation. The two networks are updated collaboratively to enhance the agent's performance in dynamic environments, aiming to maximize the expected return $J(\pi)$ as follows:

$$J(\pi) = E_{\tau \sim \pi}[\sum_{t=0}^{\infty} \gamma^t R(s_t, a_t)] \quad (2)$$

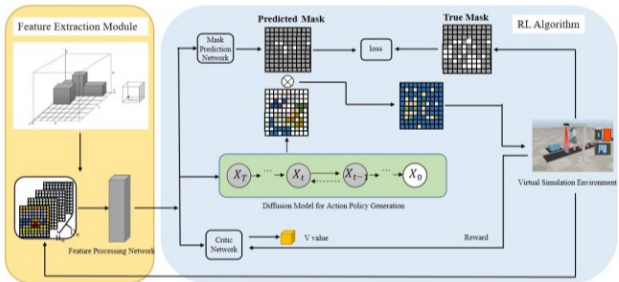

Fig. 3. Online bin packing algorithm architecture diagram.

Where $E_{\tau \sim \pi}$ denotes the expectation over trajectories τ under policy π. $\gamma$ is the discount factor, used to balance immediate rewards and future rewards, with $0 \leq \gamma < 1$. $R(s_t, a_t)$ is the immediate reward obtained by taking action $a_t$ in state $s_t$.

By incorporating this expected return formula into the description of the Actor-Critic algorithm, we clearly express the algorithm's objective and workflow in the context of 3D bin packing tasks.

To improve the model's perception of 3D spatial states, we introduce a state encoding module to transform raw packing environment states into learnable feature representations. We design a state CNN to encode the raw state vector into features. To simplify this process, we expand $d_n$ into a three-channel tensor. Each channel of $d_n$ spans an $L \times W$ matrix, with all elements being $l_n$, $w_n$ or $h_n$. Thus, the state $S_n = (H_n, D_n)$ becomes a $L \times W \times 4$ array.

In terms of reward design, we adopt a step-by-step reward mechanism. The immediate reward is based on the incremental volume occupied by the successful placement of the current object. If an action is invalid, a zero reward is given, and the current episode is terminated. This design not only increases the frequency of reward feedback but also effectively guides the agent to learn more efficient packing paths. Experiments show that, compared to using final space utilization as the terminal reward, this step-by-step reward mechanism has a stronger policy-driving effect and accelerates model convergence.

To address the dynamic changes in action feasibility in online scenarios, we design a feasibility mask prediction mechanism to avoid exploration waste caused by invalid actions. Specifically, we introduce an independent multi-layer perceptron (MLP) network. It takes the state features encoded by the CNN as input and outputs the feasibility mask for all candidate positions of the current object. During training, the mask learns from real feasibility as the supervision target. In the policy output stage, it modulates the action distribution generated by the Actor network. This ensures that the selected actions meet packing constraints such as support, boundaries, and non-overlapping.

B. *Reinforcement Learning Packing Strategy Based on Diffusion Models*

In online 3D-BPPs, the high dimensionality, multimodal distribution of the action space, and the instability of packing strategies pose significant challenges. To tackle these, this paper introduces Diffusion Probabilistic Models. These models boost the strategy modeling and action generation diversity. By simulating a gradual data denoising process, diffusion models can effectively fit complex action distribution structures. This improves the packing strategy performance in dynamic environments.

*1) Diffusion Probabilistic Models*

The data generation process in Diffusion Probabilistic Models[31] is defined as a step-wise denoising process. The structural information in data is destroyed by gradually adding noise. Then, the original structure of data is restored gradually through an inverse process. The generation process of diffusion models can be expressed as follows:

$$p_\theta(\mathbf{x}_0) = \int p_\theta(\mathbf{x}_0, \mathbf{x}_1, \ldots, \mathbf{x}_T) d\mathbf{x}_1 \ldots d\mathbf{x}_T \quad (3)$$

Here, $x_0$ represents noise-free packing action data. $x_T$ represents fully noised data. $T$ is the diffusion step. Parameters $\theta$ is optimized by minimizing the negative log-likelihood variational bound of the denoising inverse process. The inverse process is often parameterized as a Gaussian distribution with fixed time-step-related covariance. It's expressed as:

$$\mathbf{x}_{t-1} \sim N(\mathbf{x}_{t-1}; \mu_\theta(\mathbf{x}_t, t), \Sigma_\theta(\mathbf{x}_t, t)) \quad (4)$$

In packing problems, diffusion models generate reasonable packing action distributions by gradually removing noise. This process can capture the complex multi-modal distribution in the action space. Also, it can dynamically adjust the action generation strategy based on the current environmental state.

*2) Diffusion Models as Guiding Policies*

In packing tasks, traditional Gaussian policies can only fit unimodal distributions. They find it hard to adapt to the complex diversity of packing actions. Diffusion models can model any action distribution. They turn reinforcement learning problems into conditional sampling problems. Thus, they can generate optimal packing actions that are well-suited to the current environmental state, adapting to various conditions.

Specifically, let $\mathbf{y}$ be a binary random variable. It indicates the optimality of the action at time step $t$ in a trajectory. Here, $\mathbf{y} = 1$ means an optimal action, and $\mathbf{y} = 0$ a non-optimal one. By setting $\mathbf{y} = 1$, we can sample packing actions from optimal trajectories:

$$p_\theta(\mathbf{x}_0 | \mathbf{y} = 1) \propto p_\theta(\mathbf{x}_0) \cdot p_\theta(\mathbf{y} = 1 | \mathbf{x}_0) \quad (5)$$

In conditional sampling of diffusion models, it is hard to sample precisely from this distribution. But when the model is smooth enough, the reverse diffusion process can be approximated as Gaussian[31]:

$$\mu_\theta(\mathbf{x}_t, t) = \mathbf{x}_t + \sigma_t^2 \nabla_{\mathbf{x}_t} \log p_\theta(\mathbf{x}_t | \mathbf{y} = 1) \quad (6)$$

This relationship directly connects classifier guided sampling. It is used for generating class conditional images[32]

and for reinforcement learning problem settings. First, train the diffusion model on all available trajectory data. Then, train a separate model to predict the cumulative reward of trajectory samples. By adjusting the mean $\mu$ of the reverse process, the trajectory sampling can be guided to generate optimal packing actions.

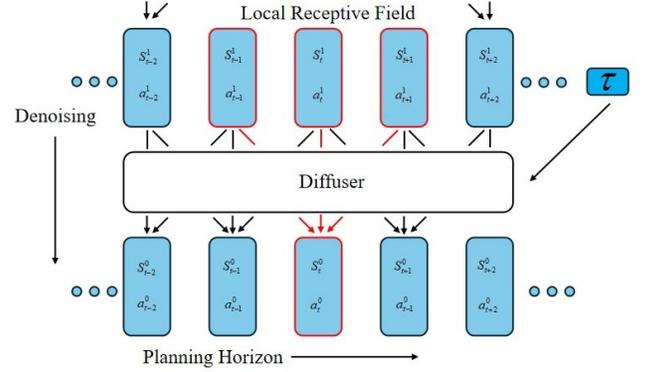

Fig. 4. Diffusion probabilistic model.

In packing tasks, the introduction of diffusion models enables the agent to dynamically generate reasonable packing action distributions. The agent uses the current packing environment's heightmap and the geometry of the object to be packed. As shown in Figure 4, through a gradual denoising process, diffusion models can capture the complex multi-modal distribution of packing actions. This improves the quality and diversity of the strategy. Experimental results show that the reinforcement learning strategy based on diffusion models is superior to traditional methods in space utilization and convergence efficiency. It shows good adaptability and generalization ability.

To realize an online 3D packing strategy based on diffusion-enhanced reinforcement learning, this paper designs a training algorithm. It combines diffusion models with the Actor-Critic framework. Table I shows the training process. It uses the diffusion model to generate action distributions and optimizes decisions with a feasibility mask. Parameters like the diffusion step $t = 100$ and learning rate are tuned to enhance space utilization.

According to the top-down algorithm logic shown in Figure 5, the online 3D bin packing algorithm consists of two main stages: the deep integrated model learning stage and the actual bin packing stage. The deep integrated model learning stage aims to enable the agent to predict the quality of current stacking actions. Considering the current cargo's dimensions and the bin packing environment's height map, the agent iteratively explores the container's current layout and determines the most favorable stacking plan. The model training strategy is as follows:

1) Cargo stacking. The agent obtains cargo online either from the current conveyor belt or from the current batch of ordered goods and loads them sequentially. During training, the focus is on accumulating each loading experience and storing it in the experience pool.

TABLE I
PSEUDOCODE FOR THE NETWORK TRAINING PROCESS

2) Random samples are selected from the experience pool for deep neural network training. With continuous updates of the cargo sequence, the deep neural network gradually acquires the ability to make decisions on current loading actions and can directly guide the agent in complex online stacking scenarios.

---

Algorithm 1: Training Process.

**Input:** Training data containing information about containers and boxes.
**Output:** Parameters of the policy network and value network.
Initialize the instance by setting up the container state $s_t$、actor parameters $\theta_a$ and critic parameters $\theta_c$;
for each $episode \in [1,2,3,...]$ do
    Sample transition batch $B = \{(S_t, a_t, r_t, S_{t+1})\} \square D_r$
    for $i = 0$ to $N / n_{age}$ do
        for $j = 0$ to $n_{age}$ do
            $\mu \leftarrow \mu(\tau^i)$ // parameters of reverse transition
            $\tau^{i-1} \square N(\mu + \alpha \sum \nabla \zeta(\mu), \Sigma^i)$ // guide using gradients of return
            $\tau^{i-1}_{s_0} \leftarrow s$ // constrain first state of plan
            Execute first action of plan $\tau^0_{a_0}$
            $S_t \leftarrow S_{t+1}, r_t \leftarrow r_{t+1}, t \leftarrow t+1$
        end for
        Calculate the loss and update $\theta_a$ and $\theta_c$;
    end for
end for

---

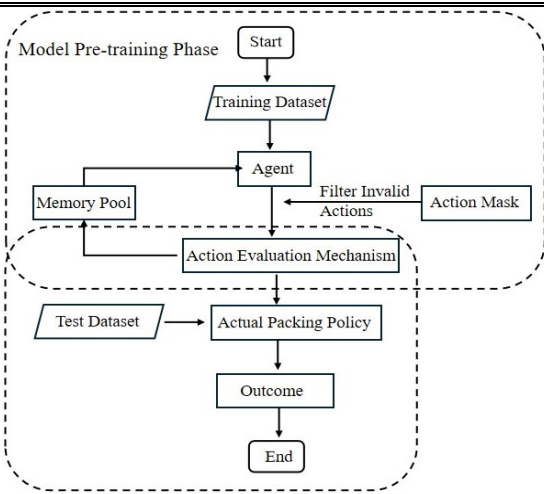

Fig.5.  Overall framework diagram of the algorithm.

## V. Experiments

The experiment was conducted on a 64-bit Ubuntu 18.04 operating system. Python programming language was used. The Pytorch open-source machine learning library was employed for development and implementation. The main hardware parameters of the platform were as follows: the CPU is a Intel(R) Xeon(R) E5-2660 v3 with 128GB of memory; the GPU was an NVIDIA GeForce RTX 3090 with 24GB of video memory.

In this experiment, the RS, CUT-1, and CUT-2 datasets proposed by Zhao[2] were used to train and test the diffusion reinforcement learning model. These datasets were designed to evaluate the performance of algorithms for the online three-dimensional BPP. The RS dataset was generated through random sampling of item sequences without specific order. It was used to test the algorithm's robustness and efficiency in random scenarios. The CUT-1 dataset was created by dividing boxes into items and sorting them by Z-coordinate. It was suitable for testing the algorithm's ability to handle structured data. The CUT-2 dataset was developed by considering stacking dependencies among items and generating sequences based on dependency order. It was closer to real-world packing scenarios and used to assess the algorithm's practicality and stability in handling complex dependencies. The container's dimensions were set as $L = W = H = 10$.

To validate the performance of the proposed method, a simulation environment was constructed. The environment simulated the online bin packing process of a robotic arm in dynamic scenarios. Figure 6 displayed the simulation scene and sensor data. These data were provided as input to the state perception module. The action distribution generated by the diffusion model was combined to achieve efficient packing. The simulation results were consistent with the space utilization of our method in Table 2. This consistency demonstrated its superiority in complex environments.

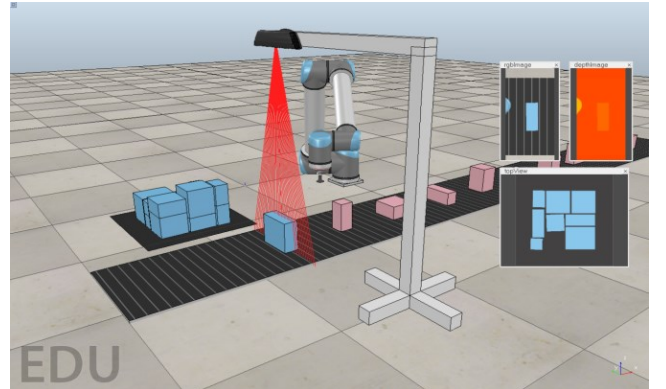

Fig. 6.  Illustrates the simulation scene for online three-dimensional bin packing.

In order to ensure effective training and stable convergence for the online three-dimensional bin packing strategy that relies on diffusion-based reinforcement learning, a suite of critical parameters had to be meticulously tuned within the reinforcement learning framework.

These parameters included the modeling of the Markov Decision Process (MDP), the design of the reward function, network optimization strategies, management of the experience replay buffer, the configuration of the diffusion model, and the overall training procedure. Table II provides a detailed breakdown of the settings and functions of these parameters, which are essential for optimizing the training process and achieving robust performance.

TABLE II
REINFORCEMENT LEARNING PARAMETER SETTINGS

| Category | Parameter | Value/Setting | Description |
|---|---|---|---|
| MDP | Discount factor γ | 0.99 | Balances immediate and future rewards |
| Reward Function | Immediate reward | Volume increase | Triggered when a package is successfully placed |
| | Penalty for invalid action | 0 | Terminates the current episode |
| Network Optimization | Actor learning rate | 1e-4 | Adam optimizer |
| | Critic learning rate | 5e-4 | Adam optimizer |
| | Batch size | 64 | Number of samples per update |
| Replay Buffer | Buffer capacity | 10000 | Stores interaction experiences |
| Diffusion Model | Diffusion steps t | 100 | Number of denoising steps |
| Diffusion Model | Training epochs | 1000 | Total number of training epochs |
| Training Settings | Max steps per epoch | 100 | Maximum interaction steps per epoch |

As shown in Figure 7, the action loss of the traditional BPP method is compared with that of the proposed diffusion model-based approach on the RS dataset. The results indicate that the proposed method achieves a rapid reduction in action loss during the early training stages and maintains a consistently lower loss thereafter. These findings suggest that the diffusion model effectively captures the complex distribution of the action space, thereby enabling the generation of more efficient action policies.

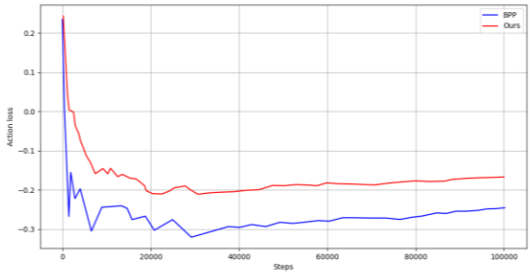

Fig. 7. Evolution of action loss over training steps

Table III compares the number of packed items and space utilization of different methods on the RS, CUT-1, and CUT-2 datasets. The proposed method based on diffusion reinforcement learning (Ours) significantly outperforms the baseline approaches. By leveraging the diffusion model's ability to represent complex multimodal action distributions, the proposed method generates high-quality packing actions that better align with the current state of the environment, thereby achieving more compact and efficient packing. Particularly on the highly random RS dataset, our approach achieves a space utilization of 57.9%, significantly surpassing the 49.7% of the BPP method, demonstrating strong generalization ability and policy robustness. Although the improvement on the RS dataset is relatively modest, this can be attributed to the dataset's high randomness, which restricts the scope of action exploration. Overall, the diffusion model improves both the adaptability and convergence speed of the method by providing structured guidance during the action generation process.

TABLE III
COMPARISON OF DIFFERENT BIN PACKING METHODS

| Method | #items / % Space uti. | | |
|---|---|---|---|
| | RS | CUT-1 | CUT-2 |
| Boundary rule (Online) | 8.7 / 34.9% | 10.8 / 41.2% | 11.1 / 40.8% |
| BPH (Online) | 8.7 / 35.4% | 13.5 / 51.9% | 13.1 / 49.2% |
| LBP (Offline) | 12.9 / 54.7% | 14.9 / 59.1% | 15.2 / 59.5% |
| BPP(Online) | 11.9 / 49.7% | 15.1 / 60.1% | 17.6 / 69.6% |
| Ours | 13.9 / 57.9% | 16.7 / 64.1% | 16.5 /63.2% |

Figure 8 illustrates the visual packing results of different methods across the RS, CUT-1, and CUT-2 datasets. The proposed approach results in more structured and compact packing arrangements, demonstrating a clear strategy of placing larger items first and heavier items at the bottom, which enhances both space utilization and stacking stability. These advantages stem from the diffusion model's capacity to model complex multimodal action distributions. By incorporating feasibility mask constraints, the model ensures both diversity and physical validity in the generated actions. In contrast to other methods that result in noticeable gaps, especially in corners and upper areas, our approach achieves a tighter packing.

These visual observations are consistent with the quantitative results in Table 3, further confirming the practical effectiveness of the diffusion model in challenging packing scenarios.

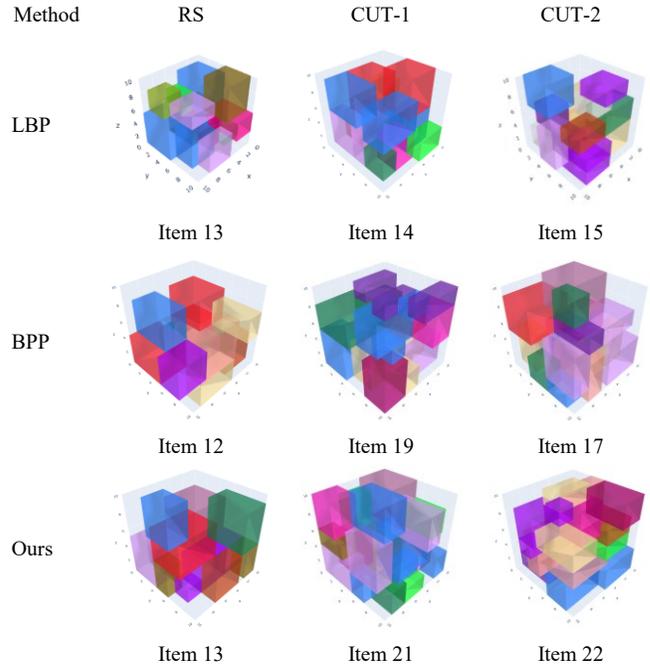

Fig. 8. Performance Comparison of Different Bin Packing Methods.

Figure 9 presents a dynamic sequence of packing steps using the proposed method on the CUT-2 dataset. From the initial state to final completion, the diffusion model progressively generates packing actions over 100 denoising steps, guided by a feasibility mask to ensure both diversity and physical validity. The final result shows a compact and orderly packing structure, consistent with the 63.2% space utilization in Table 2. However, minor height differences in subfigure (e) suggest room for further improvement. In future work, adding stability-related objectives or tuning diffusion steps will help en-hance policy precision and structural uniformity.

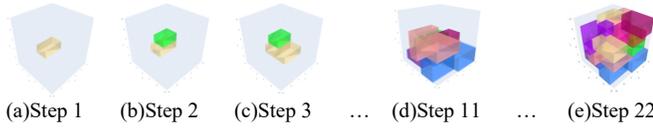

(a)Step 1　(b)Step 2　(c)Step 3　…　(d)Step 11　…　(e)Step 22

Fig. 9.　A Set of Dynamic Sequences in Online Bin Packing.

## VI. Conclusion

Compared with traditional DRL methods, our diffusion-based online 3D bin packing algorithm better models complex multimodal action distributions. Its step-by-step denoising generation improves policy diversity and robustness, allowing the agent to obtain high cumulative rewards with limited training data. Experiments verify its advantages in space utilization, convergence efficiency and generalization, especially strong adaptability in highly uncertain online scenarios with complex constraints. This work improves the applicability of diffusion reinforcement learning for combinatorial optimization, provides a robust framework for high-dimensional complex decision-making, and analyzes the collaboration between diffusion strategies and RL. It also delivers practical guidance for developing diffusion-based RL systems, promoting progress in combinatorial optimization.

## Acknowledgement

This work was supported by the Natural Science Foundation of Shandong Province under Grant ZR2024MF036.